\documentclass[11pt,a4paper]{article}
  \usepackage[hyperref]{emnlp2018}
  \usepackage{times}
  \usepackage{latexsym}
  \usepackage{amsmath, amssymb, amsthm}
  \usepackage{booktabs}
  \usepackage{graphicx}
  \usepackage{makecell}
  \usepackage{hhline}

  \usepackage{xcolor}

  \usepackage{pgfplotstable}
  \usepackage{pgfplots}
  \pgfplotsset{width=0.47\textwidth,compat=1.9, height=1.6in}

  \usepackage{url}
  \usepackage[normalem]{ulem}

\newcommand\sD{\ensuremath{\mathcal{D}}}

\newcommand\bh{\ensuremath{\mathbf{h}}}

\newcommand\bx{\ensuremath{\mathbf{x}}}
\newcommand\by{\ensuremath{\mathbf{y}}}

\newcommand\refeqn[1]{(\ref{eqn:#1})}

\newcommand\refsec[1]{Section~\ref{sec:#1}}

\newcommand\reffig[1]{Figure~\ref{fig:#1}}

\newcommand\reftab[1]{Table~\ref{tab:#1}}

\usepackage{color}

  \definecolor{lblue}{HTML}{DAE8FC}
  \definecolor{dblue}{HTML}{6C8EBF}
  \definecolor{dgreen}{HTML}{33A02C}
  \definecolor{dred}{HTML}{E31A1C}
  
  \newcommand{\dgreen}[1]{{\color{dgreen} #1}}
  \newcommand{\dred}[1]{{\color{dred} #1}}
  
  \aclfinalcopy 
  
  \title{Learning to Summarize Radiology Findings}
  
  \author{Yuhao Zhang,
  Daisy Yi Ding,
  Tianpei Qian,\\
  \textbf{Christopher D. Manning,
  Curtis P. Langlotz}
  \\
  Stanford University\\
  Stanford, CA 94305\\
  {\tt \{yuhaozhang, dingd, tianpei\}@stanford.edu}\\
  {\tt \{manning, langlotz\}@stanford.edu}\\
  \\}
  
  \date{}
  
\begin{document}
  \maketitle
  
  \begin{abstract}
The Impression section of a radiology report summarizes crucial radiology findings in natural language and plays a central role in communicating these findings to physicians.
However, the process of generating impressions by summarizing findings is time-consuming for radiologists and prone to errors.
We propose to automate the generation of radiology impressions with neural sequence-to-sequence learning.
We further propose a customized neural model for this task which learns to encode the study background information and use this information to guide the decoding process.
On a large dataset of radiology reports collected from actual hospital studies, our model outperforms existing non-neural and neural baselines under the ROUGE metrics.
In a blind experiment, a board-certified radiologist indicated that 67\% of sampled system summaries are at least as good as the corresponding human-written summaries, suggesting significant clinical validity.
To our knowledge our work represents the first attempt in this direction.
\end{abstract}
  \section{Introduction}

The radiology report documents and communicates crucial findings in a radiology study.
As shown in \reffig{motivating}, a standard radiology report usually consists of a Background section that describes the exam and patient information, a Findings section, and an Impression section \cite{kahn2009toward}.
In a typical workflow, a radiologist first dictates the detailed findings into the report, and then summarizes the salient findings into the more concise Impression section based also on the condition of the patient.

The impressions are the most significant part of a radiology report that communicate the findings.
Previous studies have shown that over 50\% of referring physicians read only the impression statements in a report \cite{lafortune1988radiological, bosmans2011radiology}.
Despite its importance, the generation of the impression statements is error-prone.
For example, crucial findings may be forgotten, which would cause significant miscommunications \cite{gershanik2011critical}.
Additionally, the process of writing the impression statements is time-consuming and highly repetitive with the dictation of the findings.
This suggests a crucial need to automate the radiology impression generation process.

\begin{figure}[!t]
  \small
  \begin{center}
  \def\arraystretch{1.5}
  \begin{tabular}{|p{0.45\textwidth}|}
  \hline
  {\bf Background:} history: swelling; pain. technique: 3 views of the \dgreen{left ankle} were acquired. comparison: no prior study available.
  \vspace{0.5em} \newline
  {\bf Findings:} there is normal mineralization and alignment. no fracture or osseous lesion is identified. the ankle mortise and hindfoot joint spaces are maintained. there is no joint effusion. the soft tissues are normal.\\
  \hhline{|=|}
  {\bf Human Impression:}\newline normal \dgreen{left ankle} radiographs.\\
  \hline
  {\bf Extractive Baseline:}\newline there is no joint effusion. \\
  \hline
  {\bf Pointer-Generator:}\newline normal \dred{right ankle}.\\ 
  \hline
  {\bf Our model:}\newline normal radiographs of the \dgreen{left ankle}.\\ 
  \hline
  \end{tabular}
  \end{center}
  \caption{An example radiology report with study background information organized into a \textbf{Background} Section, and radiology findings in a \textbf{Findings} Section. The human-written summary (or impression) and predicted summaries from different models are also shown. The extractive baseline does not summarize well, the baseline pointer-generator model generates \dred{spurious sequence}, while our model gives \dgreen{correct summary} by incorporating the background information.}
  \label{fig:motivating}
  \end{figure}

In this work, we propose to automate the generation of radiology impressions with neural sequence-to-sequence learning.
In particular, we argue that this task could be viewed as a text summarization problem, where the source sequence is the radiology findings and the target sequence the impression statements.
We collect a dataset of radiology reports from actual hospital radiographic studies, and find that this task involves both \emph{extractive summarization} where descriptions of radiology observations can be taken directly from the findings, and \emph{abstractive summarization} where new words and phrases, such as conclusions of the study, need to be generated from scratch.
We empirically evaluate existing popular summarization systems on this task and find that, while existing neural models such as the pointer-generater network can generate plausible summaries, they sometimes fail to model the study background information and thus generate spurious results.
To solve this problem, we propose a customized summarization model that properly encodes the study background information and uses the encoded information to guide the decoding process.

We show that our model outperforms existing non-neural and neural baselines on our dataset measured by the standard ROUGE metrics.
Moreover, in a blind experiment, a board-certified radiologist indicated that 67\% of sampled system summaries are at least as good as the reference summaries written by well-trained radiologists, suggesting significant clinical validity of the resulting system.
We further show through detailed analysis that our model could be reliably transferred to radiology reports from another organization, and that the model can sometimes summarize radiology studies for body parts unseen during training.

To review, our main contributions in this paper include:
(i) we propose to summarize radiology findings into impression statements with neural sequence-to-sequence learning, and to our knowledge our work represents the first attempt in this direction;
(ii) we propose a new customized summarization model to this task that improves over existing methods by better leveraging study background information;
(iii) we further show via a radiologist evaluation that the summaries generated by our model have significant clinical validity.

  \section{Related Work}

\paragraph{Early Summarization Systems.} Early work on summarization systems mainly focused on extractive approaches, where the summaries are generated by scoring and selecting sentences from the input.
\citet{luhn1958automatic} proposed to represent the input by topic words and score each sentence by the amount of topic words it contains.
\citet{kupiec1995trainable} scored sentences with a feature-based statistical classifier.
\citet{steinberger2004using} applied latent semantic analysis to cluster the topics and then select sentences that cover the most topics.
Meanwhile, various graph-based methods, such as the LexRank \cite{mihalcea2004textrank} and the TextRank algorithm \cite{erkan2004lexrank}, were applied to model sentence dependency by representing sentences as vertices and similarities as edges.
Sentences are then scored and selected via modeling of the graph properties.

\paragraph{Neural Summarization Systems.} Summarization systems based on neural network models enable abstractive summarization, where new words and phrases are generated to form the summaries.
\citet{rush2015neural} first applied an attention-based neural encoder and a neural language model decoder to this task.
\citet{nallapati2016abstractive} used recurrent neural networks for both the encoder and the decoder.
To address the limitation that neural models with a fixed vocabulary cannot handle out-of-vocabulary words,
a pointer-generator model was proposed which uses an attention mechanism that copies elements directly from the input \cite{nallapati2016abstractive, merity2016pointer, see2017get}.
\citet{see2017get} further proposed a coverage mechanism to address the repetition problem in the generated summaries.
\citet{paulus2017deep} applied reinforcement learning to summarization and more recently, \citet{chen2018fast} obtained improved result with a model that first selects sentences and then rewrites them.

\paragraph{Summarization of Radiology Reports.} Most prior work that attempts to ``summarize'' radiology reports focused on classifying and extracting information from the report text \cite{friedman1995natural, hripcsak1998extracting, elkins2000coding, hripcsak2002use}.
More recently, \citet{hassanpour2016information} studied extracting named entities from multi-institutional radiology reports using traditional feature-based classifiers. 
\citet{goff2018automated} built an NLP pipeline to identify asserted and negated disease entities in the impression section of radiology reports as a step towards report summarization. 
\citet{cornegruta2016modelling} proposed to use a recurrent neural network architecture to model radiological language in solving the medical named entity recognition and negation detection tasks on radiology reports. 
To our knowledge, our work represents the first attempt at automatic summarization of radiology findings into natural language impression statements.

  \section{Task Definition}

We now give a formal definition of the task of summarizing radiology findings.
Given a passage of findings represented as a sequence of tokens $\bx = \{ x_1, x_2, \ldots, x_N\}$, with $N$ being the length of the findings, our goal is to find a sequence of tokens $\by = \{y_1, y_2, \ldots, y_L\}$ that best summarizes the salient and clinically significant findings in $\bx$, with $L$ being an arbitrary length of the summary.%
\footnote{While the name ``impression'' is often used in clinical settings, we use ``summary'' and ``impression'' interchangably.}
Note that the mapping between $\bx$ and $\by$ can either be modeled in an unsupervised way (as done in unsupervised summarization systems), or be learned from a dataset of findings-summary pairs.

  \section{Models}

In this section we introduce our model for the task of summarizing radiology findings.
As our model builds on top of existing work on neural sequence-to-sequence learning and the pointer-generator model, we start by introducing them.

\subsection{Neural Sequence-to-Sequence Model}

At a high-level, our model implements the summarization task with an encoder-decoder architecture, where the encoder learns hidden state representations of the input, and the decoder decodes the input representations into an output sequence. 

For the encoder, we use a Bi-directional Long Short-Term Memory (Bi-LSTM) network.
Given the findings sequence $\bx = \{ x_1, x_2, \ldots, x_N\}$, we encode $\bx$ into hidden state vectors with:
\begin{equation}
\bh = \text{Bi-LSTM} (\bx),
\label{eqn:encoder}
\end{equation}
where $\bh = \{ h_1, h_2, \ldots, h_N\}$. Here $h_N$ combines the last hidden states from both directions in the encoder.

After the entire input sequence is encoded, we generate the output sequence step by step with a separate LSTM decoder. 
Formally, at the $t$-th step, given the previously generated token $y_{t-1}$ and the previous decoder state $s_{t-1}$, the decoder calculates the current state $s_t$ with:
\begin{align}
s_t = \text{LSTM}(s_{t-1}, y_{t-1}).
\label{eqn:decoder}
\end{align}
We then use $s_t$ to predict the output word.
For the initial decoder state we set $s_0 = h_N$.

The vanilla sequence-to-sequence model that uses only $s_t$ to predict the output word has a major limitation: it generates the entire output sequence based solely on a vector representation of the input (i.e., $h_N$), which may result in significant information loss.
For better decoding we therefore employ the attention mechanism \cite{bahdanau2014neural, luong2015attention}, which uses a weighted sum of all input states at every decoding step.

Given the decoder state $s_t$ and an input hidden state $h_i$, we calculate an input distribution $a^t$ as:
\begin{align}
e_i^t & = v^\top \tanh(W_h h_i + W_s s_t), \\
a^t & = \text{softmax}(e^t),
\label{eqn:attention}
\end{align}
where $W_h$, $W_s$ and $v$ are learnable parameters.%
\footnote{For clarity we leave out the bias terms in all linear layers.}
We then calculate a weighted input vector as:
\begin{align}
h_t^\ast = \sum_i{a_i^t h_i}.
\end{align}
$h_t^\ast$ encodes the salient input information that is useful at decoding step $t$.
Lastly, we obtain the output vocabulary distribution at step $t$ as:
\begin{equation}
\resizebox{.89\linewidth}{!} 
{$P(y_t | \bx, y_{<t}) = \text{softmax}(V^\prime \tanh(V [s_t; h_t^\ast])),$}
\label{eqn:vocab-distrib}
\end{equation}
where $V^\prime$ and $V$ are learnable parameters.

\begin{figure*}[t]
  \centering
  \includegraphics[width=\textwidth]{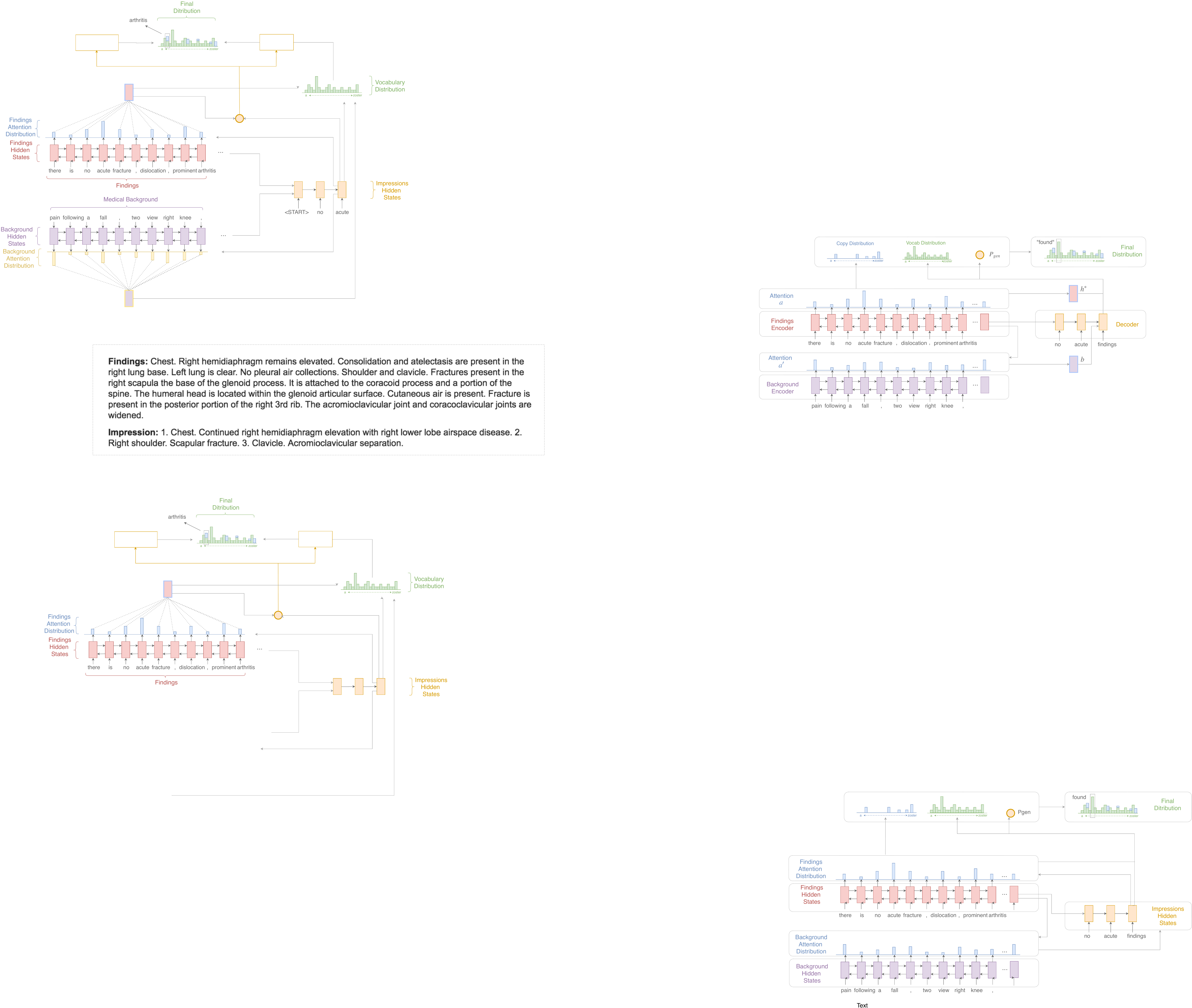}
  \caption{Overall architecture of our summarization model.}
  \label{fig:model}
\end{figure*}

\subsection{Pointer-Generator Network}
While the encoder-decoder framework described above can generate impressions from a fixed vocabulary, the model can clearly benefit from being able to ``copy'' salient observations directly from the input findings.
To add such ``copying'' capacity into the model, we use a pointer-generator network similar to the one described in \citet{see2017get}.

The main idea is that at each decoding step $t$, we allow the model to either generate a word from the vocabulary with a generation probability $p_{\text{gen}}$, or copy a word directly from the input sequence with probability $1 - p_{\text{gen}}$. We model $p_{\text{gen}}$ as:
\begin{equation}
p_{\text{gen}} = \sigma( w^\top_{h^\ast} h^\ast_t + w^\top_s s_t + w_y y_{t-1} ),
\end{equation}
where $y_{t-1}$ denotes the previous decoder output, $w_{h^\ast}$, $w_s$ and $w_y$ learnable parameters and $\sigma$ a sigmoid function.
For the copy distribution, we reuse the attention distribution $a^t$ calculated in \refeqn{attention}.
Therefore, the overall output distribution in the pointer-generator network is:
\begin{equation}
P(y_t | \bx, y_{<t}) = p_{\text{gen}} P_{\text{vocab}}(y_t) + (1 - p_{\text{gen}}) \sum_{i:x_i = y_t} a_i^t,
\end{equation}
where $P_{\text{vocab}}(y_t)$ is the same as the output distribution in \refeqn{vocab-distrib}.

\subsection{Incorporating Study Background Information}

The background part of a radiology report is also important, since crucial information such as the purpose of the study, the body part involved and the condition of the patient are often mentioned only in the background.
A straightforward way of incorporating the background information is to prepend all the background text to the findings, and treat the entire sequence as input to the pointer-generator network.
However, as we show in \refsec{results}, this naive method in fact hurts the summarization quality, presumably because the model cannot sufficiently distinguish between the findings and the background information, which as a result leads to insufficient modeling of both the findings and the background.
To solve this, we propose to encode the background text with a separate attentional encoder, and use the resulting background representation to guide the decoding process in the summarization model (\reffig{model}).

For clarity we now use $\bx^b$ to denote the background token sequence, and $\bx$ to denote the actual findings section.
Our goal is then to find $\by$ that maximizes $P(\by | \bx, \bx^b)$.
To do this, we again obtain the hidden state vectors $\bh$ of the findings section as in \refeqn{encoder}.
Similarly, we obtain the hidden state vectors of the background text with $\bx^b$ as input using a separate Bi-LSTM encoder:
\begin{equation}
\bh^b = \text{Bi-LSTM}^b (\bx^b).
\end{equation}
Next, we calculate a distribution over $\bh^b$ as:
\begin{align}
e_i^\prime & = {v^\prime}^\top \tanh(W_b h_i^b + W_h h_N), \\
a^\prime & = \text{softmax}(e^\prime),
\end{align}
where $W_b$ and $W_h$ are learnable parameters and $h_N$ the last hidden state of the findings encoder.
The distribution $a^\prime$ models the importance of tokens in the background section.
We then obtain a weighted representation of the background text as:
\begin{equation}
b = \sum_i{a_i^\prime h_i^b},
\end{equation}
where vector $b$ has the same size as $h^b$, and encodes the salient background information.

Lastly, we use the background vector $b$ to guide the decoding process, by modifying the recurrent kernel of the decoder LSTM in \refeqn{decoder} to be:
\begin{align}
\label{eqn:background-decoder}
\left[
\begin{array}{c}
i_t\\f_t\\o_t\\u_t
\end{array}
\right]
& =
\left[
\begin{array}{c}
\sigma\\\sigma\\\sigma\\\tanh{}
\end{array}
\right]
W \cdot
\left[
\begin{array}{c}
s_{t-1}\\y_{t-1}\\b
\end{array}
\right], \\
c_t & = f_t \cdot c_{t-1} + i_t \cdot u_t, \\
s_t & = o_t \cdot \tanh(c_t),
\end{align}
where $i_t$, $f_t$, $o_t$ denotes the input, forget, and output gates, $W$ the weight matrix and $c_t$ the internal cell of the LSTM repectively, and $\cdot$ represents an element-wise multiplication.
Again for clarity we leave out the bias terms in \refeqn{background-decoder}.
As a result, every state in the decoding process is directly influenced by the information encoded by the background vector $b$.
The rest of the model, including the calculation of the vocabulary distribution and the copy distribution, remains the same.

  \section{Experiments}
\label{experiments}

To test the effectiveness of our summarization model, we collected reports of radiographic studies from the picture archiving and communication system (PACS) at the Stanford Hospital.
We describe our data collection process, baseline models and experimental setup in this section, and present the results and discussions in \refsec{results}.

\subsection{Data Collection}

Reports of all radiographic studies from 2000 to 2014 were collected.
We first tokenized all reports with Stanford CoreNLP \cite{manning2014stanford}, and filtered the dataset by excluding reports where 
(1) no findings or impression section can be found;
(2) multiple findings or impression sections can be found but cannot be aligned;
or (3) the findings have fewer than 10 words or the impression has fewer than 2 words.

We removed body parts where only a small number of cases are available, and included reports of the top 12 body parts in the PACS system to maintain generalizability.
For common body parts with more than 10k reports (e.g., chest), we subsampled 10k reports from them.

\begin{figure}
  \centering
  \setlength{\abovecaptionskip}{-2pt}
	\pgfplotstableread[row sep=\\,col sep=&]{
  x & count \\
  chest & 10000 \\
  abdomen & 10000 \\
  pelvis & 10000 \\
  spine & 10000 \\
  knee & 8095 \\
  ankle & 7066 \\
  shoulder & 6900 \\
  foot & 6327 \\
  wrist & 6100 \\
  hand & 4716 \\
  elbow & 4347 \\
  tibia & 3576 \\
}\stats

\begin{tikzpicture}[font=\small]
\begin{axis}[
ybar,
yticklabel={\pgfmathprintnumber{\tick} k},
symbolic x coords={chest, abdomen, pelvis, spine, knee, ankle, shoulder, foot, wrist, hand, elbow, tibia},
xtick=data,
x tick label style={rotate=45, anchor=east},
tick align=inside,
ylabel={\# Examples},
]
\addplot[color=dblue, fill=lblue] table[x=x, y expr=\thisrow{count}/1000]{\stats};
\end{axis}
\end{tikzpicture}
  \caption{Number of examples split by body part in the collected Stanford Hospital dataset.}
  \label{fig:body-part}
\end{figure}
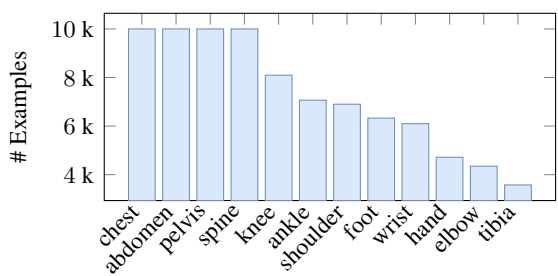

This results in a dataset with a total of 87,127 reports. We further randomly split the dataset into a 70\% training (60,990), a 10\% development (8,712) and a 20\% test set (17,425). We show the dataset statistics split by body part in \reffig{body-part}.


\begin{table*}[t]
  \centering
  \begin{tabular}{lccc}
    \toprule
    System   & ROUGE-1 & ROUGE-2 & ROUGE-L \\
    \midrule
    Extractive Baseline: S\&J-LSA & 29.39 &	16.27 &	27.38 \\
    Extractive Baseline: LexRank & 30.48 &	17.09 &	28.49 \\
    Pointer-Generator & 46.51 & 33.39 & 45.07 \\
    Pointer-Generator ($\oplus$ Background) & 45.39 & 32.60 & 44.05\\
    Our model & \bf{48.56} & \bf{35.25} & \bf{47.06} \\
    \bottomrule
  \end{tabular}
  \caption{Main results on the test set of the Stanford reports. ``$\oplus$ Background'' represents prepending the background section to the findings section to form the input to the model. All the ROUGE scores have a 95\% confidence interval of at most $\pm 0.50$ as calculated by the official ROUGE script.}
\label{tab:main-results}
\end{table*}

\subsection{Baseline Models}

For our main experiments, we compare our model against several competitive non-neural and neural systems on the collected dataset. Unless otherwise stated, the baseline models take only the findings section as input.%
\footnote{We find that when the background section is prepended to the input, the extractive baseline models may select sentences in the background part as the summary, resulting in deteriorated performance.}

\paragraph{S\&J-LSA.} This is an extractive approach described by \citet{steinberger2004using}, which applies Latent Semantic Analysis (LSA) to summarization. It first scores ``concept'' clusters by applying singular value decomposition to the term-by-sentence co-occurence matrix derived from the passage. Sentences with the top scored concepts are then kept as the summaries.

\paragraph{LexRank.} LexRank is another popular extractive model introduced by \citet{erkan2004lexrank}.
In LexRank, a passage is represented as a graph of sentences, and a connectivity matrix based on intra-sentence cosine similarity is used as the adjacency matrix of the graph. Sentences are scored by the eigenvector centrality in the graph, and the highest scored sentences are kept.

\paragraph{Pointer-Generator.} We also run the baseline pointer-generator model introduced by \citet{see2017get}.
We find the ``coverage'' mechanism described in the paper did not improve summary quality in our task and therefore did not use it for simplicity.
We compare our model with two versions of the pointer-generator model: one with only the findings section as input and another one with the background sections prepended to the findings section as input.

\subsection{Experimental Setup}

\paragraph{Evaluation Metrics.} In our main experiments we evaluate the models with the widely-used ROUGE metric \cite{lin2004rouge}.
We report the $F_1$ scores for ROUGE-1, ROUGE-2 and ROUGE-L, which measure the word-level unigram-overlap, bigram-overlap and the longest common sequence between the reference summary and the system predicted summary respectively.

\paragraph{Word Vectors.} To enable knowledge transfer from a larger corpus, we applied the GloVe algorithm \cite{pennington2014glove} to a corpus of 4.5 million radiology reports of all modalities (e.g., X-ray, CT) and body parts. We used the resulting 100-dimensional word vectors to initialize all word embedding layers in our neural models, and empirically found this to improve the performance of our neural models by about 1 ROUGE-L score.

\paragraph{Implementations \& Model Details.} For the two non-neural extractive baselines, we use their open implementations.%
\footnote{\url{https://github.com/miso-belica/sumy}}
For both of them, we select the top $N$ scored sentences to form the summary and treat $N$ as a hyperparameter. We use $N=3$ in our experiments as it yields best scores on the dev set.
We implemented all neural models with PyTorch.%
\footnote{\url{https://pytorch.org/}}
To train the neural models we append a special \texttt{<EOS>} token to the end of every reference summary.
We then employ the standard teacher-forcing with the reference summaries and optimize the negative log-likelihood loss using the Adam optimizer \cite{kingma2014adam}.
We tune all hyperparameters on the dev set.
We use 2-layer Bi-LSTM for all encoders, and set the hidden size to be 100 for each direction; 1-layer LSTM for the decoder and set the hidden size to be 200.
During inference, we employ the standard beam search with a beam size of 5.
We stop decoding whenever a \texttt{<EOS>} token is predicted, and otherwise use a maximum output sequence length of 100.

  \section{Results \& Analysis}
\label{sec:results}


\begin{figure*}[t]
  \centering
\scriptsize
\def\arraystretch{1.5}
\begin{tabular}{ | p{0.3\textwidth} | p{0.3\textwidth} | p{0.3\textwidth} | } 
\hline
{\bf Background:} radiographic examination of the abdomen. clinical history: xx years of age, male, please obtain upright and lateral decub. comparison: \dgreen{abdominal x-ray $<$date$>$}. procedure comments: two views of the abdomen.
\vspace{0.5em}
\newline
{\bf Findings:} median sternotomy wires are seen in the anterior chest wall in addition to several mediastinal clips and an aicd. trace bilateral pleural effusions are noted. interval increase in small bowel dilatation compared to previous study with multiple air-fluid levels, consistent with small bowel obstruction. there is a paucity of colonic gas. no pneumoperitoneum.
&
{\bf Background:} \dgreen{three views of the right shoulder and three views of the left shoulder}: $<$date$>$. clinical history: an xx-year-old female with bilateral shoulder pain.
\vspace{0.5em}
\newline
{\bf Findings:} three views of the right shoulder consisting of external rotation, axillary, and scapular views demonstrate no evidence of fracture or dislocation. the joint spaces are well-maintained without evidence of degenerative change. there is normal mineralization throughout. three views of the left shoulder $\ldots$ are well-maintained without evidence of degenerative change. mineralization is normal throughout.
&
{\bf Background:} three views of the abdomen: $<$date$>$. comparison: $<$date$>$. clinical history: a xx-year-old male status post hirschsprung's disease repair.
\vspace{0.5em}
\newline
{\bf Findings:} the supine, left-sided decubitus and erect two views of the abdomen show increased dilatation of the small bowel since the prior exam on $<$date$>$. there are multiple air-fluid levels, suggesting bowel obstruction. no free intraperitoneal gas is present.
\\

\hhline{|===|}
{\bf Human:} small bowel dilatation with multiple air-fluid levels and colonic decompression consistent with small bowel obstruction.
&
{\bf Human:} unremarkable radiographs of \dgreen{bilateral shoulders}. 
&
{\bf Human:} increased dilatation of the small bowel with multiple air-fluid levels, suggesting bowel obstruction. \dgreen{no free intraperitoneal gas.}
\\ 
\hline

{\bf Extractive Baseline:} median sternotomy wires are seen in the anterior chest wall in addition to several mediastinal clips and an aicd.
&
{\bf Extractive Baseline:} three views of the \dred{right shoulder} consisting of external rotation, axillary, and scapular views demonstrate no evidence of fracture or dislocation.
&
{\bf Extractive Baseline:} the supine, left sided decubitus and erect two views of the abdomen show increased dilatation of the small bowel since the prior exam on $<$data$>$.
\\ 

\hline
{\bf Pointer-Generator:} interval increase in bowel dilatation, consistent with bowel obstruction.
&
{\bf Pointer-Generator:} no evidence of fracture or dislocation of the \dred{right shoulder}.
&
{\bf Pointer-Generator:} increased dilatation of small bowel, suggesting small bowel obstruction.
\\

\hline
{\bf Our model:} interval increase in small bowel dilatation compared to \dgreen{abdominal x-ray dated $<$date$>$} with multiple air-fluid levels, consistent with small bowel obstruction.
&
{\bf Our model:} unremarkable \dgreen{bilateral shoulders}.
&
{\bf Our model:} increased dilatation of small bowel, suggesting bowel obstruction. \dgreen{no free intraperitoneal gas}.
\\ 
\hline
\end{tabular}

  \caption{Sampled test examples and system predictions from the Stanford dataset. First example: our model learns to relate the summary with a previous study mentioned only in the background section. Second: our model correctly summarizes the body part involved in the study. Third: our model correctly includes more crucial information as found in the human summary.}
  \label{fig:main-examples}
\end{figure*}

\subsection{Main Results}

We present results of our main experiments in \reftab{main-results}.
We find that the two non-neural extractive models perform comparably, and both are able to obtain non-trivial subsequence overlap with the reference summaries as measured by ROUGE scores.
However, a baseline neural pointer-generator that combines the sequence generation and the copy mechanism beats the non-neural baselines substantially on all metrics.
We confirm that naively incorporating the study background information by prepending the background section directly to the input findings in the pointer-generator model in fact hurts the performance (noted by $\oplus$ Background).
In comparison, our model benefits from using the separately encoded background vector to guide the decoding process, and achieves best scores on all ROUGE metrics.

We also present sampled test examples and system output in \reffig{main-examples}.
We find that compared to the non-neural extractive baselines, the neural models are not limited by sentences in the findings section and therefore generate summaries of better quality.
For example, the neural models learn to compose the summary by combining observation phrases from different sentences, or by generating new conclusive phrases such as ``negative study''.
Compared to the pointer-generator model, our model learns to correctly utilize relevant background information (e.g., previous study or exam information) to improve the summary.

\subsection{Clinical Validity with Radiologist Evaluation}

One potential shortcoming of the ROUGE metrics is that they only measure the similarity between the predicted summary and the reference summary, but do not sufficiently reflect the overall grammaticality or utility of the predictions.
Therefore, we also conducted evaluations with a board-certified radiologist to understand the clinical validity of our system generated summaries.

In this evaluation, we randomly sampled 100 examples from our test set. We ran our best model over these 100 examples, and presented each example along with the corresponding system predicted summary and reference human-written summary to the radiologist. 
We randomly ordered the predicted and reference summary such that the correspondence cannot be guessed from the order. 
The radiologist was asked to select which of the two summaries was better, or that they have roughly equal quality.

\begin{table}[t]
  \centering
  \begin{tabular}{lc}
    \toprule
    Category   & Percentage \\
    \midrule
    Human Summary Wins &	33 \\
    System Prediction Wins & 16 \\
    Roughly Equal Quality & 51 \\
    \bottomrule
  \end{tabular}
  \caption{Radiologist evaluation result on 100 sampled test examples. For a total of 67 examples, the radiologist indicated that the system summary is at least as good as the human-written summary.}
\label{tab:human-eval}
\end{table}

\reftab{human-eval} presents the result. 
For 51 examples, the radiologist indicated that the human-written and system-generated summaries are equivalent. 
For 16 examples, the radiologist preferred the system summary, and for the remaining 33 examples, the radiologist preferred the human-written summary. 
Note that under our setting, a randomly generated sequence would have almost zero chance to be indicated as good as the human-written summary. 
We therefore believe the result suggests significant clinical validity of our system.

\begin{table}[t]
  \centering
  \setlength{\tabcolsep}{0.35em}
  \begin{tabular}{lccc}
    \toprule
    System   & ROUGE-1 & ROUGE-2 & ROUGE-L \\
    \midrule
    LexRank & 15.42 &	\phantom{0}5.65 &	14.60 \\
    Our model & 35.02 & 20.79 & 34.56 \\
    \bottomrule
  \end{tabular}
  \caption{Cross-organization evaluation results on the Indiana University chest x-ray dataset. All the ROUGE scores have a 95\% confidence interval of at most $\pm 1.10$ as calculated by the official ROUGE script.}
\label{tab:cross-org}
\end{table}

\subsection{Does the model transfer to reports from another organization?}

Deploying a clinical NLP system at an organization different from the one where the training data comes from is a common need.
However, this is challenging in that medical practitioners including radiologists from different organizations tend to go through different training and follow different templates or styles when writing medical text.
Here we aim to understand the cross-organization transferability of our summarization model. 

We use the publicly available Indiana University Chest X-ray Dataset \cite{demner2015preparing}, which consists of chest X-ray images paired with the corresponding radiology reports.
We filtered the reports with the same set of rules and arrived at a collection of 2,691 unique reports.
We used this dataset as the test set, and ran our best model trained on our own dataset directly on it.
The results are shown in \reftab{cross-org} and sampled examples are shown in the first two columns of \reffig{other-examples}.
We find that our model again outperforms the baseline extractive model substantially in this transfer setting, and the generated summaries are both grammatical and clinically meaningful.

\begin{figure*}[t]
  \centering
\scriptsize
\def\arraystretch{1.5}
\begin{tabular}{ | p{0.3\textwidth} | p{0.3\textwidth} | p{0.3\textwidth} | } 
\hline
Cross-organization & Cross-organization & Cross-body part: Knee \\
\hline
{\bf Background:} indication: xxxx year old male with end-stage renal disease on hemodialysis
\vspace{0.5em}
\newline
{\bf Findings:} the heart size is mildly enlarged. there is tortuosity of the thoracic aorta. no focal airspace consolidation, pleural effusions or pneumothorax. no acute bony abnormalities. &

{\bf Background:} indication: xxxx year old female, hypoxia. comparison: pa lateral views of the chest dated xxxx.
\vspace{0.5em}
\newline
{\bf Findings:} bilateral emphysematous again noted and lower lobe fibrotic changes. postsurgical changes of the chest including cabg procedure, stable. stable valve artifact. there are no focal areas of consolidation. no large pleural effusions. no evidence of pneumothorax. $\ldots$
contour abnormality of the posterior aspect of the right 7th rib again noted, stable. &

{\bf Background:} radiographic examination of the knee: $<$date$>$ $<$time$>$. clinical history: xx-year-old man with right knee pain. comparison: none. procedure comments: 2 views of the right knee were performed.
\vspace{0.5em}
\newline
{\bf Findings:} there is no visible fracture or malalignment. likely small joint effusion. mild fullness in the popliteal region of the right knee may represent a baker 's cyst. mild soft tissue swelling along the medial aspect of the knee is present.\\
\hhline{|===|}
{\bf Human:} cardiomegaly without acute pulmonary findings.&
{\bf Human:} no acute cardiopulmonary abnormality. stable bilateral emphysematous and lower lobe fibrotic changes.&
{\bf Human:} no acute bony abnormality. likely joint effusion and soft tissue swelling along the medial aspect of the knee. 
\\ 
\hline 
{\bf Our model:} mild cardiomegaly. no radiographic evidence of acute cardiopulmonary process.&
{\bf Our model:} stable postsurgical changes of the chest as described above. no evidence of pneumothorax.&
{\bf Our model:} mild soft tissue swelling along the medial aspect of the knee. no fracture or malalignment.\\
\hline
\end{tabular}

  \caption{First two columns: sampled examples from the Indiana University dataset and system output in the cross-organization evaluation. Last column: sampled test example of a ``knee'' study in our cross-body part evaluation. }
  \label{fig:other-examples}
\end{figure*}

\begin{table}[t]
  \centering
  \setlength{\tabcolsep}{0.4em}
  \begin{tabular}{lccc}
    \toprule
    Body Part & ROUGE-1 & ROUGE-2 & ROUGE-L \\
    \midrule
    Chest & 31.24 &	17.99 &	30.38 \\
    Abdomen & 28.90 & 17.23 & 27.83 \\
    Knee & 48.78 & 35.07 & 47.49 \\
    \bottomrule
  \end{tabular}
  \caption{Cross-body part evaluation results of our neural model on the Stanford dataset. All the ROUGE scores have a 95\% confidence interval of at most $\pm 0.75$ as calculated by the official ROUGE script.}
\label{tab:cross-body-part}
\end{table}

\subsection{Does the model transfer to body parts unseen during training?}

Radiology studies conducted on different body parts often include vastly different observations and diagnosis.
For example, while ``lung base opacity'' is a common observation in chest radiographic studies, it does not exist in musculoskeletal studies.
In practice, an organization may not have adequate report data that covers some rare body parts.
It is therefore interesting to test to what extent our summarization model can generalize to reports for body parts unseen during training.

We study this by simulating the condition where a specific body part is not present in the training data. 
Given the entire dataset $\sD$, and a subset of the dataset $\sD_B$ that corresponds to a body part $B$, we reserved the entire subset $\sD_B$ as test data, and used $\sD - \sD_B$ for training (90\%) and validation (10\%).
\reftab{cross-body-part} presents the evaluation results for body part ``chest'', ``abdomen'' and ``knee''.
We find that for ``chest'' and ``abdomen'', the system summaries degrade substantially when the corresponding data were not seen during training.
However, the predicted summaries degrade less for ``knee'' when reports of it were not seen during training, presumably because the model can learn to summarize reasonably well from reports of other close musculoskeletal studies such as ``ankle'' or ``elbow'' studies.
We confirm this by examining the model predictions: in the example shown in the last column of \reffig{other-examples}, the model learns to compose the summary with salient observations such as ``tissue swelling'' and ``fracture'', while being able to copy the anatomy ``knee'' (unseen during training) from the findings section.

\subsection{What is the model missing on?}

\begin{table}[t]
  \centering
  \begin{tabular}{lc}
    \toprule
    Category & Percentage \\
    \midrule
    Good Summary &	63 \\
    \midrule
    Missing Critical Info. & 24 \\
    Inaccurate/Spurious Info. & 8 \\
    Redundant & 4 \\
    Ungrammatical & 6 \\
    \bottomrule
  \end{tabular}
  \caption{Error analysis on 100 sampled dev examples from the Stanford dataset.}
\label{tab:error-analysis}
\end{table}

\begin{figure*}[t]
  \centering
	\scriptsize
\def\arraystretch{1.5}
\begin{tabular}{ | p{0.3\textwidth} | p{0.3\textwidth} | p{0.3\textwidth} | } 
\hline
Error type: missing critical information & Error type: redundant summary & Error type: ungrammatical summary \\
\hline
{\bf Background:} radiographic examination of the lumbar spine: $<$time$>$. clinical history: $<$age$>$, lower back pain. comparison: none. procedure comments: 4 views of the lumbar spine.
\vspace{0.5em}
\newline
{\bf Findings:} five non-rib bearing lumbar type vertebral bodies are present. there is trace retrolisthesis of l5 on s1. there is no evidence of instability on flexion and extension views. the spinal alignment is otherwise normal. the disc spaces and vertebral body heights are preserved. there is no visible fracture. no visible facet joint arthropathy or pars defects.&

{\bf Background:} radiographic examination of the shoulder: $<$time$>$. clinical history: $<$age$>$ years of age, pain in joint involving shoulder region. comparison: outside study dated $<$date$>$. procedure comments: single axillary view of the left shoulder.
\vspace{0.5em}
\newline
{\bf Findings:} single axillary view of the shoulder again demonstrates a highly comminuted fracture of the humeral head and likely fracture of the scapular body. the humeral head appears located on the glenoid.&

{\bf Background:} radiographic examination of the shoulder: $<$time$>$. clinical history: $<$age$>$ years of age, xray exam of lower spine 2 or 3 views. x-ray exam of right shoulder complete. comparison: none. procedure comments: three views of the right shoulder.
\vspace{0.5em}
\newline
{\bf Findings:} a calcification of the rotator cuff is seen above the greater tuberosity. there is no fracture or malalignment. the soft tissues and visualized lung are unremarkable.\\
\hhline{|===|}
{\bf Human:} \dred{trace retrolisthesis of l5 on s1 with no evidence of instability with motion.} otherwise normal lumbar spine.&
{\bf Human:} redemonstration of a highly comminuted fracture of the humeral head and likely fracture of the scapular body . the humeral head appears to be located on the glenoid .&
{\bf Human:} no acute bony or joint abnormality, but there is calcification of the rotator cuff that may be due to calcific tendinitis.
\\
\hline 
{\bf Our model:} no acute bony or articular abnormality.&
{\bf Our model:} highly comminuted \dred{fracture of the scapular body} and likely \dred{fracture of the scapular body}.&
{\bf Our model:} \dred{calcification} acute bony or joint abnormality.\\
\hline
\end{tabular}

  \caption{Examples of different types of errors that our system makes on the Standord dataset. Words that are missing from or are erroneously included in the model predictions are highlighted in red.}
  \label{fig:error-example}
\end{figure*}

Lastly, we run a detailed error analysis on 100 sampled dev examples.
We focus on four types of errors:
(1) missing critical information, if the predicted summary fails to include some clinically important information;
(2) inaccuate/spurious information, if the predicted summary contains observations or conclusions that are inaccurate, or that do not exist in the findings;
(3) redundant summary, if the predicted summary is repetitive or over-verbose;
and (4) ungrammatical summary, if the predicted summary contains significant grammatical errors.
For each example, we examine whether it contains any of the errors by comparing it with the reference summary; otherwise we classify it as a good summary.
Note that an example can be assigned to more than one error categories.

We include examples of different error types in \reffig{error-example}, and present the result of error analysis in \reftab{error-analysis}.
We find that 63\% examples are qualitatively close to the reference summary, which aligns well with the radiologist evaluation result.
Among the four error categories, missing critical information is the most common error with 24\% examples, suggesting that the summaries may be improved with explicit modeling of the importance of different radiology findings.
We also find through qualitative analysis that the model tends to miss on followup procedures recommended by the human radiologist, since these procedures are often not included in the findings section and generating them needs significant understanding of the study and domain knowledge.

  \section{Conclusion}

In this paper we proposed to generate radiology impressions from findings via neural sequence-to-sequence learning.
We proposed a customized neural model for this task which uses encoded background information to guide the decoding process.
We collected a dataset from actual hospital studies and showed that our model not only outperforms non-neural and neural baselines, but also generates summaries with significant clinical validity and cross-organization transferability.

  \section*{Acknowledgments}

We thank Peng Qi and the anonymous reviewers for their helpful suggestions.
  
  \bibliography{main}
  \bibliographystyle{acl_natbib_nourl}
  
  \end{document}